\documentclass{article}
\usepackage{spconf,epsfig}
\usepackage{graphicx}
\usepackage{amsmath,amssymb,mathtools}

\DeclareMathOperator\erf{erf}
\let\OLDthebibliography\thebibliography
\renewcommand\thebibliography[1]{
  \OLDthebibliography{#1}
  \setlength{\parskip}{0pt}
  \setlength{\itemsep}{0pt plus 0.3ex}
}

\begin{document}\sloppy

\def\x{{\mathbf x}}
\def\L{{\cal L}}

\title{Boundary Uncertainty in a Single-Stage Temporal Action Localization Network}
%
\name{Tingting Xie \qquad Christos Tzelepis \qquad Ioannis Patras}
\address{School of Electronic Engineering and Computer Science,
Queen Mary University of London}
%
\maketitle
\begin{abstract}
In this paper, we address the problem of temporal action localization with a single stage neural network. In the proposed architecture we model the boundary predictions as uni-variate Gaussian distributions in order to model their uncertainties, which is the first in this area to the best of our knowledge. We use two uncertainty-aware boundary regression losses: first, the Kullback-Leibler divergence between the ground truth location of the boundary and the Gaussian modeling the prediction of the boundary and second, the expectation of the $\ell_1$ loss under the same Gaussian. We show that with both uncertainty modeling approaches improve the detection performance by more than $1.5\%$ in mAP@tIoU=0.5 and that the proposed simple one-stage network performs closely to more complex one and two stage networks. 
\end{abstract}
%

\section{Introduction}\label{sec:intro}


Driven by the need to process a large number of untrimmed videos generated daily by various video capturing devices, temporal action localization is drawing increasing attention from the research community~\cite{yeung2016end,shou2016temporal,gao2017turn,zhao2017temporal,lin2017single,lin2018bsn,xie2019exploring,lin2019bmn,zeng2019graph,lin2019fast,liu2019multi}. 


Temporal action localization typically involves first, generating video segments as candidate action proposals, and second, jointly classifying them into an action class and regressing/refining their temporal boundaries so as to better localize them in time~\cite{gao2017cascaded,lin2017single,chao2018rethinking,xie2019exploring}. However, for actions in the wild, that is in unconstrained scenarios, there are large variations in how actions are performed -- this makes it difficult to predict accurate boundaries. Also, unlike object boundaries, there might even be no sensible definition of what the exact temporal extent of a action is. This makes temporal boundary annotations subjective and, possibly, not consistent across different persons. Such issues are not taken into consideration by traditional regression losses used for boundary refinement (such as $\ell_1$ loss~\cite{gao2017turn,gao2017cascaded}).

To address the above issues, inspired by recent works (e.g.,~\cite{he2019bounding,kingma2013vae}), we firstly propose to model the boundary predictions as uni-variate Gaussian distributions in temporal action localization, for which we learn their means and variances -- the latter express the uncertainty about each prediction. Then, we exploit this kind of uncertainty by using two uncertainty-aware boundary regression losses. First, we use the Kullback-Leibler (KL) loss between a dirac, representing the ground truth location of the boundary and the uni-variate Guassian -- this, is the cost proposed in~\cite{he2019bounding} for the problem of object detection. Second, we propose to approximate the expectation of $\ell_1$ loss, that is typically used as regression loss -- to back-propagate the error with respect to the parameters of the Gaussian, resort to the reparametrization trick and an approximation by sampling as in \cite{kingma2013vae}.

Experimental evaluation of the above losses shows that the network learns to assign large variances to the samples that are predicted to be far from the ground truth boundary values. As the network converges and the predictions become more accurate, this behaviour changes and the network assigns small variances to accurate predictions. Both uncertainty-aware losses improve detection and localization performance.

\begin{figure*}[ht] 
    \centering
    \includegraphics[width=\linewidth]{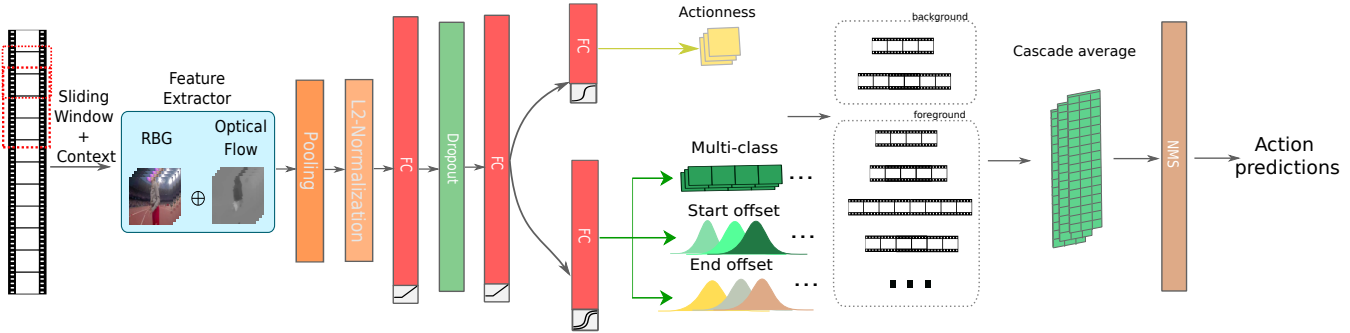}
    \caption{Network architecture. Given an untrimmed video, sliding window proposals are generated firstly, and unit-level features are extracted by feature extractor, and then, they are passed to the network. The networks mainly constitute by two branches, one (upper branch) outputs actionness (binary) classification score to indicate if this proposal is an action; while the other one (lower branch) outputs the classification score for each class and the corresponding regression offsets (distribution) to the start and end time. During testing, temporal boundaries are adjusted in a cascaded way by feeding the refined windows back to the system for further refinement. All parameters in each cascade step are shared.}
\label{fig:ssdlike_arch}
\end{figure*}

The contributions of the paper are summarized as follows:
\begin{enumerate}
    \item We propose a simple and effective one-stage network that introduces and exploits uncertainty modeling of the boundary location for temporal action localisation. To the best of our knowledge, this is the first paper that that does so in this domain.
    \item For action localization we propose to use two uncertainty-aware losses: the first, based on the KL-divergence to model the difference between distributions and the second, based on the expectation of the $\ell_1$ loss proposed by us. 
    \item We show that the uncertainty modeling improves over the adopted baseline, and that our one-stage network achieves comparable results with recent one- and two-stage networks on THUMOS'14.\footnote{Code will be made public here: https://github.com/}
\end{enumerate}


\section{Related work}\label{sec:related_work}

\textbf{One-stage action localization detectors:} Single-stage networks have been extensively used for detection~\cite{lin2017single,xu2017r,buch2017end,huang2019decoupling}. However, their performance is usually inferior than that of two-stage networks. The SSD-like detection architecture presented in~\cite{lin2017single} seems to perform well, but temporal span modeling of videos has more variations and more arbitrary compared to spatial information in objects. Thus, it is hard to use hand-designed anchor to cover them all to get the accurate boundary without explicitly modeling the temporal information, especially for actions with large duration. \cite{xu2017r} is the first to propose an end-to-end network, but C3D feature~\cite{le2011learning} has been proved to be inferior than two-stream feature~\cite{simonyan2014two} used in~\cite{gao2017turn, xie2019exploring}. \cite{buch2017end} exploits C3D as a feature extractor and GRU~\cite{cho2014gru}, a concise and elegant way to model temporal information and make predictions of the offset. However, experimental results show that GRU is not sufficient in order to learn representation for accurate localization compared to CNN-based methods. In this paper, we try to alleviate the drawback of the above methods using our one-stage network.

\textbf{Uncertainty Learning in DNNs:} To improve robustness and interpretability of discriminant Deep Neural Networks (DNNs), introducing and learning under uncertainty is receiving increasing attention among the research community~\cite{gal2016dropout,kendall2017uncertainties,he2019bounding,shi2019pfe}. In this respect, two main categories of uncertainty are studied: \textit{model uncertainty} and \textit{data uncertainty}. Model uncertainty refers to the uncertainty of the model parameters given the training data and can be reduced by collecting additional data~\cite{gal2016dropout}. Data uncertainty accounts for the uncertainty in output whose primary source is the inherent noise in input data~\cite{kendall2017uncertainties,he2019bounding,shi2019pfe}. Despite the fact that a few methods have been proposed for dealing with data uncertainty in classification and regression problems (e.g., in segmentation~\cite{kendall2017uncertainties} or object detection~\cite{he2019bounding} tasks), to the best of our knowledge, this the first work that does that in the temporal action localization domain.

\section{Method}\label{sec:method}
\subsection{Baseline architecture}\label{subsec:architect}
    
In this work, we propose a simple single-stage network, which will serve as our baseline architecture. The proposed network draws inspiration from the standard two-stage approach that includes a proposal generation stage and a detection stage. Both of them could be considered as standard classification-regression networks that take as input a fixed-size feature representation scheme extracted from temporal clips of varying lengths. First, in the proposal generation stage a binary classification classifies the segment as being background or foreground (i.e., being one of a set of known actions). Second, in the detection stage, a coupled regression-classification scheme refines the segment boundaries and classifies it into one of the knows action classes. In this paper, we combine the two stages together into a singe-stage network that conducts end-to-end action detection and localization by having two branches to perform binary classification and multi-classification/regression separately, as shown in Fig.~\ref{fig:ssdlike_arch}.
    
Specifically, for each input proposal, to partially preserve input temporal structure, we divide each input proposal into $k$ parts and apply average pooling to each part, as in~\cite{xie2019exploring}, to get a fixed-dimensional feature representation scheme, and then a $\ell_2$-normalization layer and a fully-connected layer (along with a ReLU layer) are followed. After that, there are two branches of fully-connected layer: the first branch is only doing binary classification, which indicates if this proposal is an action or not; while the second branch is to output multi-classification scores and the refined start, and end offsets corresponding to refinements of the boundaries for each action category. Different from the lower branch shown in Fig.~\ref{fig:ssdlike_arch}, the baseline network only predicts start offsets and end offsets with $\ell_1$ regression loss; the distribution prediction will be described in detail in Sect. \ref{subsec:uam}.

\subsection{Classification}

Before discussing the different boundary regression methods, let us define the training set with supervision as follows:

$$
     \mathcal{X}=\left\{\left(\mathbf{x}^{(i)}, t_{a}^{(i)}, t_{c}^{(i)}, t_{s}^{(i)}, t_{e}^{(i)}\right)\right\}_{i=1}^{N},
$$
           
where $\mathbf{x}^{(i)}$, $t_{a}^{(i)}$, $t_{c}^{(i)}$, $t_{s}^{(i)}$, and $t_{e}^{(i)}$ denote the feature vector, the actioness label, class label, the start, and the end offsets of the $i$-th training sample, respectively. $t_{a}$ is binary, which indicates that a training example depicts a class or the background. $t_{c}$ is a multi-class label, indicates the category a training example belongs to. Given a feature vector $\mathbf{x}$, the baseline network infers the actioness score $y_{a}$, multi-class score $y_{c}$, the start offset $y_{s}$, and the end offset $y_{e}$.
            
For the binary classification task, i.e., for learning the actioness score, we use the standard binary cross entropy loss. However, due to the fact that proposals that actually depict an action are far fewer than the ones that depict background, the dataset is imbalanced. To deal with this, we adopt a popular technique (see e.g.~\cite{liu2016ssd}), namely hard-negative mining, where we keep the ratio between positive and negative (with respect to actioness) samples fixed and equal to $\lambda$. For our experiments, we set $\lambda=\frac{1}{3}$. Then, the total binary loss is given as:

\begin{equation}
     \mathcal{L}_{bin}=\frac{-1}{\lvert I_p\rvert + \lvert I_n^{\lambda}\rvert}\left(\sum_{i \in I_p}\log\left(y_{a}^{(i)}\right)+\sum_{i \in I_n^{\lambda}}\log\left(1-y_{a}^{(i)}\right)\right)
\end{equation}

where $I_p$ and $I_n^{\lambda}$ denote the sets of indices of the positive and chosen negative samples, respectively.
    
\subsection{Standard multi-class classification and boundary regression}

For the multi-class classification task, suppose $N_c$ is the number of action category in the dataset, the classification loss is written by:

\begin{equation}
    \mathcal{L}_{cls} = \frac{1}{\lvert I_p\rvert}\sum_{i\in I_p} \sum_{j}^{N_c}\left(  -t_{c}^{(ij)}\log \hat{y}_c^{(ij)} \right)
\end{equation}

where

$$
    \hat{y}_c^j = \frac{\exp(y_c^j)}{\sum_j \exp(y_c^j)}
$$

For the regression task, i.e., for adjusting the start and the end offsets, typically $\ell_1$ loss is used:
        
\begin{equation}
    \mathcal{L}_{reg} = \frac{1}{\lvert I_p\rvert}\sum_{i\in I_p}\left(\lvert t_{s}^{(i)} - y_{s}^{(i)}\rvert + \lvert t_{e}^{(i)} - y_{e}^{(i)}\rvert\right)
\end{equation}
    
\subsection{Uncertainty-aware boundary regression}\label{subsec:uam}
        
As discussed above, for modeling output uncertainty, we propose to model the boundary offsets as uni-variate Gaussian distributions for which their first- and second-order moments are learned by the network (see Fig.~\ref{fig:ssdlike_arch}). That is, instead of predicting a deterministic pair of start/end boundaries, we predict a pair of uni-variate Gaussians. In the next two sections, we discuss two regression losses that exploit this kind of uncertainty; i.e., one that explicitly uses the distributions for computing the boundary regression loss and one that samples for them to approximate the expectation of $\ell_1$ loss.

\begin{figure*}[ht]
            \centering
            \begin{minipage}[b]{0.45\linewidth}
                \centering
                \includegraphics[width=\linewidth]{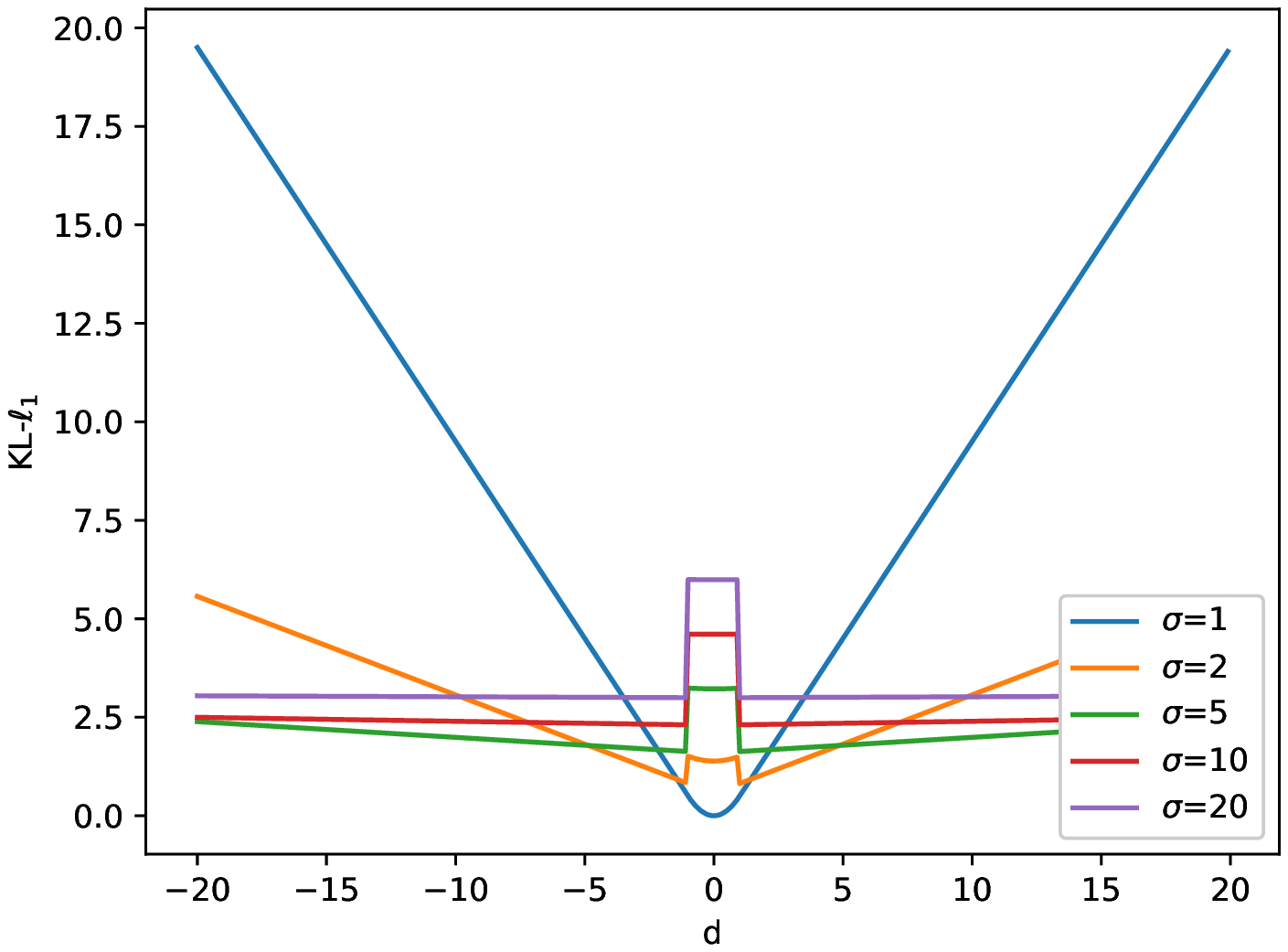}
                \label{fig:kl_div_loss}
                \centerline{(a) KL-$\ell_1$ regression loss.}\medskip
            \end{minipage}
            %
            \begin{minipage}[b]{0.45\linewidth}
              \centering
                \includegraphics[width=\linewidth]{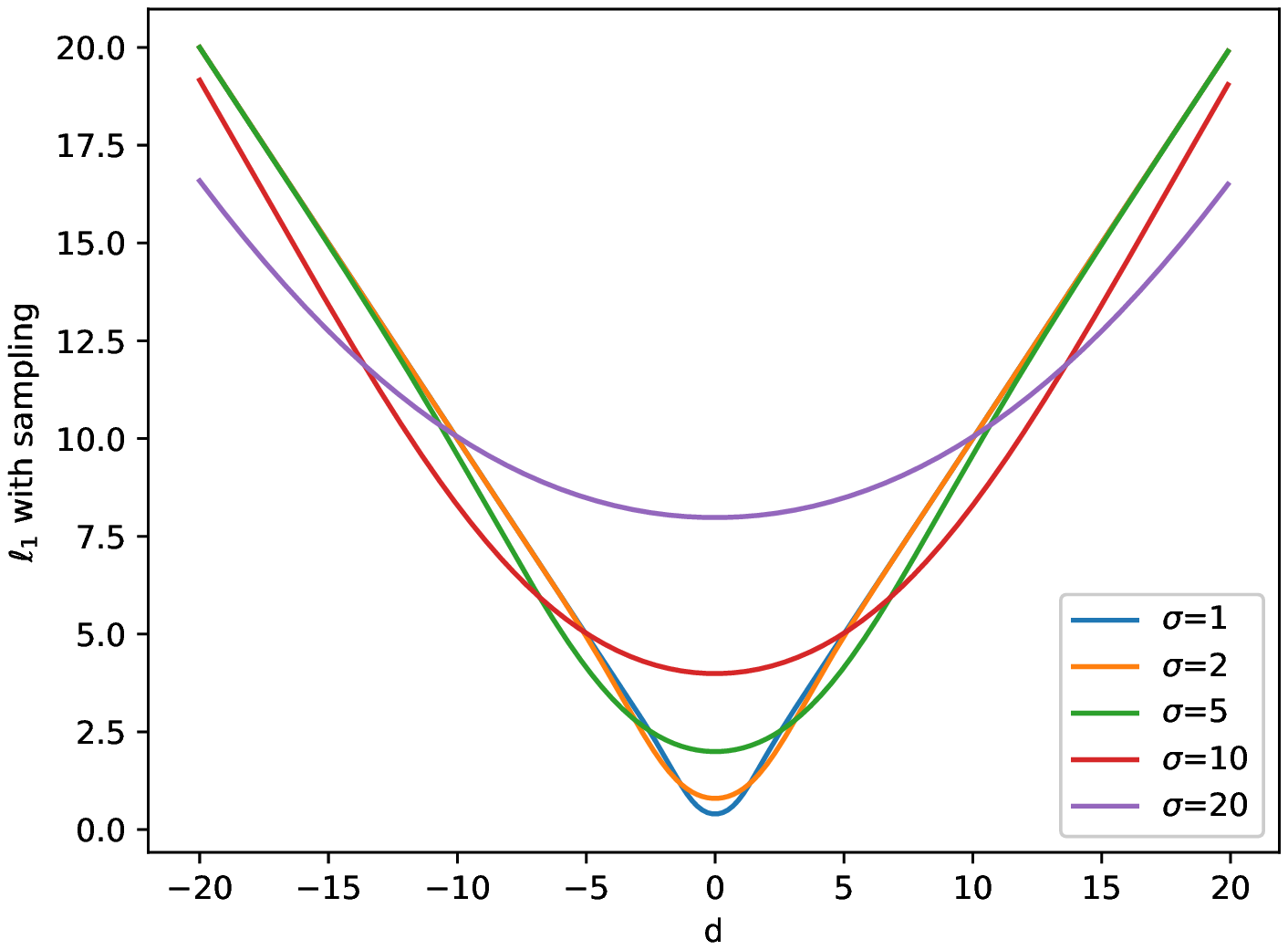}
                \label{fig:l1_loss_sigma}
              \centerline{(b) $\ell_1$ regression loss with sampling.}\medskip
            \end{minipage}
            \caption{Uncertainty-aware boundary regression losses.}
            \label{fig:fomula_loss}
\end{figure*}
    
\textbf{KL-$\ell_1$ regression loss:} Following similar arguments as in~\cite{he2019bounding}, we adopt the Kullback-Leibler divergence combined with another loss which is similar to smooth $\ell_1$ loss for computing the boundary regression loss. To this end, we treat ground truth values as Dirac delta distributions, i.e., centred at the given values, in order to indicate the lack of any prior-knowledge about their uncertainty. For the sake of simplicity, if $t$ is the ground truth value of a boundary offset, and $\mu$, $\sigma^2$ are the mean and the variance of the corresponding network's prediction, then, for $d=t-\mu$, the following regression loss is introduced when $\lvert d\rvert>1$:

\begin{equation}
    \mathcal{L}_{kl-\ell_1} = \frac{d^2}{2\sigma^2} + \frac{log(\sigma^2)}{2} + \frac{log(2\pi)}{2}, 
\end{equation}

and the modified smooth $\ell_1$ loss when $\lvert d\rvert\leq1$:

\begin{equation}
    \mathcal{L}_{kl-\ell_1} = \frac{1}{\sigma^2}\left(\lvert d\rvert - \frac{1}{2}\right) + \log(\sigma)
\end{equation}

We show the above regression loss in Fig.~\ref{fig:fomula_loss} (a). It is worth noting that for large values of $d$, i.e., for predicted offsets that are far from the corresponding ground truth values, loss is decreases for predictions with large variances. That is, using KL-$\ell_1$ loss will force the network to predict offsets with large variances in order to converge quickly. By doing this, the network is given more freedom to discard some noisy training samples by enlarging the variances of the output. On the other hand, when the network starts to converge, i.e., when the distance between the predicted offsets and the ground truth values becomes smaller than a certain threshold, the network is trying to make the variances smaller to be accurate.
    
\textbf{$\ell_1$ regression loss with sampling:} We propose an alternative uncertainty-aware boundary regression loss in order to avoid the explicit use of distributions in loss computation. In particular, at each training iteration, we sample from the predicted boundary offset distributions and compute the standard $\ell_1$ loss. In this way, we approximate the expectation of $\ell_1$ loss during training.
            
More specifically, if $t$ is the ground truth value of a boundary offset, and $\mu$, $\sigma^2$ are the mean and the variance of the corresponding network's prediction, at each iteration we sample from $\mathcal{N}(\mu,\sigma)$ and compute the $\ell_1$ loss, i.e., the quantity $\lvert t-y\rvert$. However, since the sampling operation is not a well-defined differentiable operation, and thus would render back-propagation impossible, we use the well-known reparameterization trick~\cite{kingma2013vae}. That is, by choosing one source of randomness like the uni-variate standard Gaussian $\mathcal{N}(0,1)$, we express the boundary offset prediction as:

$$
    y=\mu+\sigma\epsilon, \quad \epsilon\sim\mathcal{N}(0,1).
$$

Thus, the regression loss could be represented by (where $d = t-\mu$):

\begin{equation}
\label{equ:l_samp}
    \mathcal{L}_{samp}
    = \lvert t - \mu -\sigma\epsilon\rvert
    = |d - \sigma \epsilon| 
\end{equation}
                
In this way, we approximate during training the expected $\mathcal{L}_{samp}$, which can be analytically be expressed as follows:

\begin{equation}
    \mathbb{E}\left[\mathcal{L}_{samp}\right]
    = 
    \mathbb{E}\left[\lvert d-\sigma\epsilon\rvert\right]
    =
    d\erf\left(\frac{d}{\sqrt{2\sigma^2}}\right) + \frac{\sigma\exp\left(-\frac{d^2}{\sigma^2}\right)}{\sqrt{2\pi}}
\end{equation}

We show the expected $\ell_1$ loss in Fig.~\ref{fig:fomula_loss}(b). Compared with $\mathcal{L}_{kl-\ell_1}$, it doesn't have such a big tendency to predict offsets with big variances when $|d|$ is big. From the curve, it can be informed vaguely that $\mathcal{L}_{samp}$ tend to optimize $d$ firstly, and then turn to variances. Detailed derivation can be found in the appendix.

\section{Experiments}\label{sec:exp}

\textbf{Dataset} We evaluate the proposed methods on the popular THUMOS'14~\cite{THUMOS14} dataset, which contains 200 validation and 213 testing untrimmed videos, temporally annotated with 20 action classes. Following the standard practice~\cite{chao2018rethinking,lin2018bsn,lin2019bmn}, we train our models on the validation set and evaluate them on the testing set.

\textbf{Implementation details} Our baseline method is illustrated in Fig.~\ref{fig:ssdlike_arch}. Input dimensionality of the features that feed the first FC layer is $k\cdot d$, where $k$ is a user-defined hyper-parameter discussed in Sect.~\ref{subsec:architect} and $d=4096$ is the feature dimension of units (each unit consists of 16 frames) of the input proposal. The first FC layer has 1000 hidden units that feed the second FC layer, which outputs two branches. The first one predicts the actioness score (whether the input proposal depicts an action or background), and the second one predicts classification and regression scores. The output of this branch is a $C\times 3$ matrix in the baseline case, and a $C\times 6$ matrix in the case where both means and variances are predicted ($C=20$ denotes the number of classes). During training, we used a batch size of 128, and a rate of $10^{-3}$.

\subsection{One- vs two-stage networks}

To demonstrate the usability of our one-stage network, we compare to a similar two-stage architecture~\cite{xie2019exploring}, for which we use proposals generated from our network. The two-stage network is constituted by a proposal generation network, and a detection network, while these two has the same classification-regression structure with ours respectively. In Table~\ref{tab:onestage} we show that the proposed one-stage network achieves comparable results by involving class-agnostic along with category information in a single-stage network, with approximately half of the parameters. 

\begin{table}[t]
    \begin{center}
    \caption{Two-stage and one-stage (baseline) networks performance comparison in THUMOS'14.} 
    \label{tab:onestage}
    \begin{tabular}{c|ccccc}
      \hline
      mAP@IoU (\%) & 0.3 & 0.4 & 0.5 & 0.6 & 0.7
      \\
      \hline
      Two-stage & 49.68 & 44.67 & 36.48 & 24.29 & 13.59 \\
      One-stage & 49.46 & 44.89 & 36.22 & 25.56 & 14.98 \\
      \hline
    \end{tabular}
    \end{center}
\end{table}
    
\begin{figure*}[ht]
    \centering
    \begin{minipage}[b]{0.45\linewidth}
        \centering
        \includegraphics[width=\linewidth]{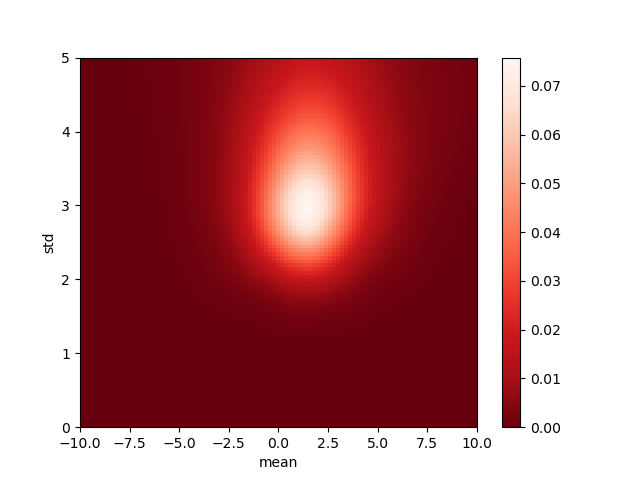}
        \centerline{(a) KL-$\ell_1$ boundary regression loss.}\medskip
    \end{minipage}
    \begin{minipage}[b]{0.45\linewidth}
      \centering
    \includegraphics[width=\linewidth]{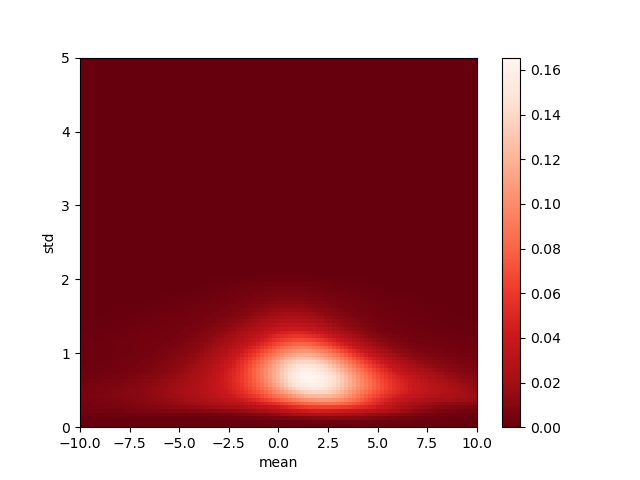}
      \centerline{(b) $\ell_1$ regression loss with sampling.}\medskip
    \end{minipage}
    \caption{Distribution of boundary offsets means and variances using uncertainty-aware boundary regression losses.}
    \label{fig:heatmap}
\end{figure*}

\subsection{Uncertainty-Aware losses}

In this section, we compare the uncertainty-aware KL-$\ell_1$ and the expected $\ell_1$-sampling boundary regression losses on THUMOS'14 dataset for the problem of temporal action detection. In Fig.~\ref{fig:heatmap}, we visualize the optimal means and variances of the offsets learned after training with the above two losses. Moreover, in Table~\ref{tab:exp_results_soa} we report the performance of the two networks. We observe that, compared to $\ell_1$-sampling loss, KL-$\ell_1$ loss encourages learning larger variances. As we discussed in Sect.~\ref{sec:method}, the network can learn more from ``easy'' samples, and ignore the ``hard'' ones by increasing their variances to enhance the detection performance, which boosts the baseline in all tIoUs by approximately $2\%$.

While for $\ell_1$-sampling loss the variances look smaller compared to KL-$\ell_1$ loss (see Fig.~\ref{fig:heatmap}(b)), it is constrained dynamically when the $\ell_1$-norm between ground truth and prediction is becoming smaller. It boosts the performance by around $1\%$ (see Table~\ref{tab:exp_results_soa}) for tIoUs apart from 0.7 by constraining the uncertainty in a relatively low level.

\textbf{KL-$\ell_1$ vs $\ell_1$-sampling regression loss} Using KL-$\ell_1$ boundary regression loss arrives at slightly better results than using $\ell_1$-sampling loss. We argue that, due to the extreme imbalance between positive and negative proposals generated by sliding window, it's more urgent to suppress the negative noisy samples rather than boost the positive boundary prediction accuracy. While KL divergence could suppress the negative proposals by enlarging the corresponding variances; $\ell_1$-sampling could give more realistic variances by boosting the regression accuracy, which leads to that. 

\subsection{Comparison to state-of-the-art}

In Table~\ref{tab:exp_results_soa} we report the experimental results of two uncertainty-aware losses compared to several related works. We note that KL-$\ell_1$ loss achieves second highest performance among the single-stage methods, even comparable with current two-stage methods; and highlight that with the uncertainty estimation our result outperforms the other one-stage methods apart from \cite{huang2019decoupling} in a large margin by more than $5\%$ in all tIoUs without bells and whistles. As to~\cite{huang2019decoupling}, the main stream branch uses~\cite{lin2017single} as the backbone network (they improve the backbone mAP@tIoU=0.5 from $24.6\%$ to $31.2\%$, which is still not as good as ours, $37.9\%$), but it has two extra branches to deal with. That is, a proposal generation and a classification branch need to be trained as well, which triples the parameters of~\cite{lin2017single} to achieve the reported performance.

\begin{table}[h]
\caption{Temporal action localization methods on THUMOS'14 with various tIoU thresholds.}
    \small
	\begin{center}
	    \begin{tabular}{lllllll}
	    \hline
	    Method & \multicolumn{5}{c}{mAP@IoU (\%)} \\ 
        & 0.3 & 0.4 & 0.5 & 0.6 & 0.7 \\ \hline
        \multicolumn{6}{c}{Two-stage methods} \\ \hline
        {CDC ~\cite{shou2017cdc}}               & 40.1 & 29.4 & 23.3 & 13.1 & 7.9   \\
        {SSN ~\cite{zhao2017temporal}}          & 51.9 & 41.0 & 29.8 & --   & --    \\
        {CBR ~\cite{gao2017cascaded}}     & 50.1 & 41.3 & 31.0 & 19.1 & 9.9   \\
        {Faster rcnn~\cite{chao2018rethinking}} & 53.2 & \textbf{48.5} & \textbf{42.8} &  \textbf{33.8} & \textbf{20.8}  \\ 
        {BSN ~\cite{lin2018bsn}}          & 53.5 & 45.0 & 36.9 & 28.4 & 20.0  \\
        {TAD ~\cite{xie2019exploring}} & 52.5 & 46.6 & 37.4 & 24.5 & 12.4\\
        {TBN ~\cite{zhang2019boundary}}   & 53.8 & 47.1 & 39.1 & 29.7 & 20.8  \\
        {BMN ~\cite{lin2019bmn}} & 56.0 & 47.4 & 38.8 & 29.7 & 20.5 \\
        {GTAN ~\cite{long2019gaussian}} & \textbf{57.8} & 47.2 & 38.8 & - & - \\
        \hline 
        \multicolumn{6}{c}{One-stage methods} \\ \hline
        {RL ~\cite{yeung2016end}}        & 36.0 & 26.4 & 17.1 & --   & --    \\
        {SSAD ~\cite{lin2017single}}     & 43.0 & 35.0 & 24.6 & 15.4 & 7.7   \\
        {SAP ~\cite{huang2018sap}}  & - & - & 27.7 & - & -\\
        {SS-TAD ~\cite{buch2017end}}     & 45.7 & - & 29.2 & - & 9.6 \\
        {Decoup-ssad ~\cite{huang2019decoupling}}   & \textbf{60.2} & \textbf{54.1} & \textbf{44.2} & \textbf{32.3} & \textbf{19.1}  \\  \hline 
        \multicolumn{6}{c}{Ours} \\ \hline
        Baseline   & 49.5 & 44.9 & 36.2 & 25.6 & 15.0 \\ 
        $\mathbb{E}[\ell_1$] (sampling)    & 50.5 & 45.1 & 37.7 & 26.1 & 14.9 \\
        KL-$\ell_1$ & \textbf{51.8} & \textbf{47.7} & \textbf{37.9} & \textbf{27.6} & \textbf{16.0} \\
        \hline
        \end{tabular}
        
    \label{tab:exp_results_soa}
    \end{center}
\end{table}

\section{Conclusion}\label{sec:conclusion}
In this paper we propose an uncertainty-aware boundary regression loss for the problem of temporal action localization in videos. We model boundary offset predictions as uni-variate Gaussian distributions and we compute the expectation of $\ell_1$ loss for improving localization. We compare with another uncertainty-aware loss that explicitly uses the predicted distributions, which we apply to the problem of temporal action localization for the first time. In the future, we intend to investigate the use of the predicted variances in the test phase in the direction of improving inference.

\bibliographystyle{IEEEbib}
\bibliography{icme2020main}

\begin{thebibliography}{10}

\bibitem{yeung2016end}
Serena Yeung, Olga Russakovsky, Greg Mori, and Li~Fei-Fei,
\newblock ``End-to-end learning of action detection from frame glimpses in
  videos,''
\newblock in {\em CVPR}, 2016, pp. 2678--2687.

\bibitem{shou2016temporal}
Zheng Shou, Dongang Wang, and Shih-Fu Chang,
\newblock ``Temporal action localization in untrimmed videos via multi-stage
  cnns,''
\newblock in {\em CVPR}, 2016, pp. 1049--1058.

\bibitem{gao2017turn}
Jiyang Gao, Zhenheng Yang, Chen Sun, Kan Chen, and Ram Nevatia,
\newblock ``Turn tap: Temporal unit regression network for temporal action
  proposals,''
\newblock {\em arXiv preprint arXiv:1703.06189}, 2017.

\bibitem{zhao2017temporal}
Yue Zhao, Yuanjun Xiong, Limin Wang, Zhirong Wu, Xiaoou Tang, and Dahua Lin,
\newblock ``Temporal action detection with structured segment networks,''
\newblock in {\em ICCV}, 2017, vol.~8.

\bibitem{lin2017single}
Tianwei Lin, Xu~Zhao, and Zheng Shou,
\newblock ``Single shot temporal action detection,''
\newblock in {\em ACM Multimedia}. ACM, 2017, pp. 988--996.

\bibitem{lin2018bsn}
Tianwei Lin, Xu~Zhao, Haisheng Su, Chongjing Wang, and Ming Yang,
\newblock ``Bsn: Boundary sensitive network for temporal action proposal
  generation,''
\newblock {\em ECCV}, 2018.

\bibitem{xie2019exploring}
Tingting Xie, Xiaoshan Yang, Tianzhu Zhang, Changsheng Xu, and Ioannis Patras,
\newblock ``Exploring feature representation and training strategies in
  temporal action localization,''
\newblock {\em ICIP}, 2019.

\bibitem{lin2019bmn}
Tianwei Lin, Xiao Liu, Xin Li, Errui Ding, and Shilei Wen,
\newblock ``Bmn: Boundary-matching network for temporal action proposal
  generation,''
\newblock {\em ICCV}, 2019.

\bibitem{zeng2019graph}
Runhao Zeng, Wenbing Huang, Mingkui Tan, Yu~Rong, Peilin Zhao, Junzhou Huang,
  and Chuang Gan,
\newblock ``Graph convolutional networks for temporal action localization,''
\newblock in {\em ICCV}, 2019, pp. 7094--7103.

\bibitem{lin2019fast}
Chuming Lin, Jian Li, Yabiao Wang, Ying Tai, Donghao Luo, Zhipeng Cui, Chengjie
  Wang, Jilin Li, Feiyue Huang, and Rongrong Ji,
\newblock ``Fast learning of temporal action proposal via dense boundary
  generator,''
\newblock {\em arXiv preprint arXiv:1911.04127}, 2019.

\bibitem{liu2019multi}
Yuan Liu, Lin Ma, Yifeng Zhang, Wei Liu, and Shih-Fu Chang,
\newblock ``Multi-granularity generator for temporal action proposal,''
\newblock in {\em CVPR}, 2019, pp. 3604--3613.

\bibitem{gao2017cascaded}
Jiyang Gao, Zhenheng Yang, and Ram Nevatia,
\newblock ``Cascaded boundary regression for temporal action detection,''
\newblock {\em arXiv preprint arXiv:1705.01180}, 2017.

\bibitem{chao2018rethinking}
Yu-Wei Chao, Sudheendra Vijayanarasimhan, Bryan Seybold, David~A Ross, Jia
  Deng, and Rahul Sukthankar,
\newblock ``Rethinking the faster r-cnn architecture for temporal action
  localization,''
\newblock in {\em CVPR}, 2018, pp. 1130--1139.

\bibitem{he2019bounding}
Yihui He, Chenchen Zhu, Jianren Wang, Marios Savvides, and Xiangyu Zhang,
\newblock ``Bounding box regression with uncertainty for accurate object
  detection,''
\newblock in {\em CVPR}, 2019, pp. 2888--2897.

\bibitem{kingma2013vae}
Diederik~P Kingma and Max Welling,
\newblock ``Auto-encoding variational bayes,''
\newblock {\em arXiv preprint arXiv:1312.6114}, 2013.

\bibitem{xu2017r}
Huijuan Xu, Abir Das, and Kate Saenko,
\newblock ``R-c3d: Region convolutional 3d network for temporal activity
  detection,''
\newblock in {\em ICCV}, 2017, vol.~6, p.~8.

\bibitem{buch2017end}
S~Buch, V~Escorcia, B~Ghanem, L~Fei-Fei, and JC~Niebles,
\newblock ``End-to-end, single-stream temporal action detection in untrimmed
  videos,''
\newblock in {\em BMVC}, 2017.

\bibitem{huang2019decoupling}
Yupan Huang, Qi~Dai, and Yutong Lu,
\newblock ``Decoupling localization and classification in single shot temporal
  action detection,''
\newblock {\em arXiv preprint arXiv:1904.07442}, 2019.

\bibitem{le2011learning}
Quoc~V Le, Will~Y Zou, Serena~Y Yeung, and Andrew~Y Ng,
\newblock ``Learning hierarchical invariant spatio-temporal features for action
  recognition with independent subspace analysis,''
\newblock in {\em CVPR}. IEEE, 2011, pp. 3361--3368.

\bibitem{simonyan2014two}
Karen Simonyan and Andrew Zisserman,
\newblock ``Two-stream convolutional networks for action recognition in
  videos,''
\newblock in {\em NIPS}, 2014, pp. 568--576.

\bibitem{cho2014gru}
Kyunghyun Cho, Bart Van~Merri{\"e}nboer, Caglar Gulcehre, Dzmitry Bahdanau,
  Fethi Bougares, Holger Schwenk, and Yoshua Bengio,
\newblock ``Learning phrase representations using rnn encoder-decoder for
  statistical machine translation,''
\newblock {\em arXiv preprint arXiv:1406.1078}, 2014.

\bibitem{gal2016dropout}
Yarin Gal and Zoubin Ghahramani,
\newblock ``Dropout as a bayesian approximation: Representing model uncertainty
  in deep learning,''
\newblock in {\em ICML}, 2016, pp. 1050--1059.

\bibitem{kendall2017uncertainties}
Alex Kendall and Yarin Gal,
\newblock ``What uncertainties do we need in bayesian deep learning for
  computer vision?,''
\newblock in {\em NIPS}, 2017, pp. 5574--5584.

\bibitem{shi2019pfe}
Yichun Shi, Anil~K Jain, and Nathan~D Kalka,
\newblock ``Probabilistic face embeddings,''
\newblock {\em arXiv preprint arXiv:1904.09658}, 2019.

\bibitem{liu2016ssd}
Wei Liu, Dragomir Anguelov, Dumitru Erhan, Christian Szegedy, Scott Reed,
  Cheng-Yang Fu, and Alexander~C Berg,
\newblock ``Ssd: Single shot multibox detector,''
\newblock in {\em ECCV}. Springer, 2016, pp. 21--37.

\bibitem{THUMOS14}
Y.-G. Jiang, J.~Liu, A.~Roshan~Zamir, G.~Toderici, I.~Laptev, M.~Shah, and
  R.~Sukthankar,
\newblock ``{THUMOS} challenge: Action recognition with a large number of
  classes,'' http://crcv.ucf.edu/THUMOS14/, 2014.

\bibitem{shou2017cdc}
Zheng Shou, Jonathan Chan, Alireza Zareian, Kazuyuki Miyazawa, and Shih-Fu
  Chang,
\newblock ``Cdc: convolutional-de-convolutional networks for precise temporal
  action localization in untrimmed videos,''
\newblock in {\em CVPR}. IEEE, 2017, pp. 1417--1426.

\bibitem{zhang2019boundary}
Tao Zhang, Shan Liu, Thomas Li, and Ge~Li,
\newblock ``Boundary information matters more: Accurate temporal action
  detection with temporal boundary network,''
\newblock in {\em ICASSP}. IEEE, 2019, pp. 1642--1646.

\bibitem{long2019gaussian}
Fuchen Long, Ting Yao, Zhaofan Qiu, Xinmei Tian, Jiebo Luo, and Tao Mei,
\newblock ``Gaussian temporal awareness networks for action localization,''
\newblock in {\em CVPR}, 2019, pp. 344--353.

\bibitem{huang2018sap}
Jingjia Huang, Nannan Li, Tao Zhang, Ge~Li, Tiejun Huang, and Wen Gao,
\newblock ``Sap: Self-adaptive proposal model for temporal action detection
  based on reinforcement learning,''
\newblock in {\em AAAI}, 2018.

\end{thebibliography}

\cleardoublepage

\addtocontents{toc}{\protect\contentsline{chapter}{Appendix:}{}}
\section{Detailed Derivation of the expectation of $\ell_1$ loss}

\textbf{Lemma. 1} Suppose the predicted offset is y, and the corresponding ground truth is $t$, the $\ell_1$ loss is defined by:

$$\mathcal{L} = |t - y|$$

where $y=\mu+\sigma\epsilon, \quad \epsilon\sim\mathcal{N}(0,1)$. The expectation of $\ell_1$ loss can be analytically expressed as follows:
$$
                \mathbb{E}\left[\mathcal{L}\right]
                =
                d\erf\left(\frac{d}{\sqrt{2\sigma^2}}\right) + \frac{\sigma\exp\left(-\frac{d^2}{\sigma^2}\right)}{\sqrt{2\pi}},
$$

\noindent \textit{Proof.} 

According to Equ. \ref{equ:l_samp}, 
$$
    \mathcal{L} = \lvert t - \mu -\sigma\epsilon\rvert = |d - \sigma \epsilon|, \quad d = t-\mu
$$

Then,

\begin{align}
    \mathcal{L} &= \left\{\begin{array}{ll}d - \sigma \epsilon, \quad if \quad  d-\sigma \epsilon \geq 0 \\
    \sigma \epsilon-d, \quad if \quad d-\sigma \epsilon < 0
    \end{array}
    \right. \\
    &= \left\{\begin{array}{ll}
    d - \sigma \epsilon, \quad if \quad \epsilon \leq \frac{d}{\sigma}\\
    \sigma \epsilon-d, \quad if \quad \epsilon > \frac{d}{\sigma}
    \end{array}
    \right.  
\end{align}  

Thus,

\begin{align}
    \mathbb{E}\left[\mathcal{L}\right] &= \int_\mathbb{R} |d - \sigma \epsilon| p(\epsilon) d\epsilon \\
    &=\int_{-\infty}^{d/\sigma} (d - \sigma \epsilon)p(\epsilon) d\epsilon + \int_{d/\sigma}^{+\infty} (\sigma \epsilon-d)p(\epsilon) d\epsilon    
\end{align}

where $p(\epsilon)$ is the Probability density function of $\epsilon$. Thus, we will know: $\int_\mathbb{R} p(\epsilon) d\epsilon = 1$, that is to say:

\begin{equation}
\label{equ:p_epsilon_int}
\int_{-\infty}^{d/\sigma} p(\epsilon)d\epsilon + \int_{d/\epsilon}^{+\infty} p(\epsilon)d\epsilon = 1  
\end{equation}

$\mathbb{E}\left[\mathcal{L}\right]$ could be written into:

\begin{align}
\label{equ:prefinal}
\mathbb{E}\left[\mathcal{L}\right] =  d\int_{-\infty}^{d/\sigma} p(\epsilon) d\epsilon - \sigma \int_{-\infty}^{d/\sigma} \epsilon p(\epsilon) d\epsilon \\
+ \sigma \int_{d/\sigma}^{+\infty} \epsilon  p(\epsilon) d\epsilon - d \int_{d/\sigma}^{+\infty} p(\epsilon) d\epsilon
\end{align}

At this stage, we are going to divide $\mathbb{E}\left[\mathcal{L}\right]$ into parts and conquer one by one. Based on Equation \ref{equ:p_epsilon_int}, we can derive:

\begin{align}
 \int_{d/\epsilon}^{+\infty} p(\epsilon) d\epsilon = 1 - \int_{-\infty}^{d/\infty} p(\epsilon) d\epsilon   
\end{align}

Then:

\begin{align}
 - d \int_{d/\sigma}^{+\infty} p(\epsilon) d\epsilon = 
 - d + d\int_{-\infty}^{d/\infty} p(\epsilon) d\epsilon   
\end{align}

As $\epsilon\sim\mathcal{N}(0,1)$, we know $\int_\mathbb{R} \epsilon p(\epsilon) d\epsilon = 0$, thus:

\begin{equation}
\label{equ:part14}
    \int_{-\infty}^{d/\sigma} \epsilon p(\epsilon) d\epsilon + \int_{d/\sigma}^{+\infty} \epsilon p(\epsilon) d\epsilon = 0
\end{equation}

Thus:

\begin{equation}
\label{equ:part23}
 \sigma \int_{d/\sigma}^{+\infty} \epsilon p(\epsilon) d\epsilon = -\sigma \int_{-\infty}^{d/\sigma} \epsilon p(\epsilon) d\epsilon
\end{equation}

Finally, based on Equation  \ref{equ:prefinal}, \ref{equ:part14} and \ref{equ:part23}, we can derive $\mathbb{E}\left[\mathcal{L}\right]$ into:

\begin{align}
\mathbb{E}\left[\mathcal{L}\right] &= 2d \int_{-\infty}^{d/\sigma} p(\epsilon) d\epsilon - d -2\sigma \int_{-\infty}^{d/\sigma} \epsilon p(\epsilon) d\epsilon \\
&= 2d \Phi(d/\sigma) -d + 2\sigma \frac{\exp(-(d/\sigma)^2)}{2\sqrt{2\pi}}\\
&= d \left[2 \Phi(d/\sigma) -1 \right] + \sigma \frac{\exp(-(d/\sigma)^2)}{\sqrt{2\pi}}\\
&= d \left[2 \frac{1}{2}(1+\erf(d/\sqrt{2\sigma^2})) -1 \right] + \sigma \frac{exp(-(d/\sigma)^2)}{\sqrt{2\pi}}\\
&= d \erf(d/\sqrt{2\sigma^2})) + \sigma \frac{exp(-(d/\sigma)^2)}{\sqrt{2\pi}}
\end{align}

\end{document}